\documentclass[11pt,a4paper]{article}
\usepackage{amssymb}
\usepackage[hyperref]{acl2020}
\usepackage{times}
\usepackage{latexsym}
\usepackage{url}

\usepackage{natbib}
\usepackage{adjustbox}
\usepackage{tabularx} 
\usepackage{booktabs} 
\usepackage[margin=1in]{geometry} 
\usepackage{lipsum} 
\usepackage{adjustbox}
\usepackage{verbatim}
\usepackage{listings}
\usepackage{booktabs} 
\usepackage{xcolor} 
\usepackage{float}
\usepackage{graphicx}
\usepackage{algpseudocode}
\usepackage{algorithm}
\floatname{algorithm}{Reward function}

\usepackage{amsmath}

\usepackage[vietnamese]{babel}

\definecolor{tableheader}{rgb}{0.2,0.2,0.7}
\definecolor{exampletext}{rgb}{0.5,0.5,0.5}

\usepackage{microtype}

\aclfinalcopy 

\title{GreenMind: A Next-Generation Vietnamese Large Language Model for Structured and Logical Reasoning}

\author{
  Luu Quy Tung\textsuperscript{1} \quad
  Hoang Quoc Viet\textsuperscript{1}\thanks{Corresponding author: \texttt{viethq5@greennode.ai}} \quad
  Pham Bao Loc\textsuperscript{1} \quad
  Vo Trong Thu\textsuperscript{2}\\
  \textsuperscript{1}GreenNode.ai \quad 
  \textsuperscript{2}John Von Neumann Institute\\
  \{\texttt{tunglq},\,\texttt{viethq5},\texttt{locpb}\}@\texttt{greennode.ai}, 
  \texttt{thuvt@jvn.edu.vn}
}

\begin{document}
\maketitle

\begin{abstract}
Chain-of-Thought (CoT) is a robust approach for tackling LLM tasks that require intermediate reasoning steps prior to generating a final answer. In this paper, we present \textbf{GreenMind-Medium-14B-R1} \footnote{\href{https://huggingface.co/GreenNode/GreenMind-Medium-14B-R1}{https://huggingface.co/GreenNode/GreenMind-Medium-14B-R1}}, the Vietnamese reasoning model inspired by the finetuning strategy based on Group Relative Policy Optimization. We also leverage a high-quality Vietnamese synthesized reasoning dataset and design two reward functions to tackle the main limitations of this technique: i) Language mixing, where we explicitly detect the presence of biased language characters during the process of sampling tokens, and ii) We leverage Sentence Transformer-based models to ensure that the generated reasoning content maintain factual correctness and do not distort the final output. Experimental results on the Vietnamese dataset from the VLSP 2023 Challenge demonstrate that our model outperforms prior works and enhances linguistic consistency in its responses. Furthermore, we extend our evaluation to SeaExam — a multilingual mutiple-choices dataset, showing the effectiveness of our reasoning method compared to few-shot prompting techniques.
\end{abstract}

\section{Introduction}
The rapid advancement of Large Language Models (LLMs) has transformed the approach to handling complex tasks such as question answering and multiple-choice problems. Many open-source LLMs have demonstrated impressive capabilities in natural language understanding.
However, for tasks like question answering and multiple-choice reasoning,  the act of users prompting models to produce direct answers only often fails to ensure accuracy. Meanwhile, at each generation step, models rely on the probability distribution over a list of candidate tokens to select the potential one by greedy or random sampling algorithms. Consequently, producing only a short sequence of tokens as the final output does not guarantee correctness, as these distributions are conditioned solely on the preceding input tokens. This implies that the models often lack the contextual understanding necessary for reasoning toward a correct answer. To address this issue, the CoT \cite{wei2022chain} technique remains an effective approach to fully leverage the power of next token prediction. CoT encourages the model to articulate a sequence of intermediate reasoning steps, which facilitates the resolution of tasks that require multi-step logical thinking. To further enhance the reasoning capabilities of language models, a series of reinforcement learning-based methods have been proposed. Reinforcement Learning with Human Feedback (RLHF) \cite{ouyang2022training} leveraged human-provided feedback to refine LLM outputs, ensuring that the reasoning steps generated by CoT align more closely with human-like judgment and reasoning. Proximal Policy Optimization (PPO) balanced exploration and exploitation by updating the reasoning policy using a clipped objective function, which helps avoid large, destabilizing changes while enhancing CoT reasoning across multiple steps. 

In this study, we introduce \textbf{GreenMind-Medium-14B-R1}, a fine-tuned LLM model capable of reasoning for tasks within the Vietnamese community. Our model leverages the GRPO technique \cite{shao2402deepseekmath, guo2025deepseek}, which has been shown to enhance reasoning effectiveness for the CoT method as well as reduce computational costs. However, the limitation of this approach is its inability to control for linguistic bias (typically English and Chinese) inherent in the base models, which means that generated responses may contain characters from the language with the dominant training dataset. Additionally, the quality control of the reasoning process has not been addressed in the original work \cite{guo2025deepseek}, which may lead to content distortion relative to the original query. To tackle these challenges, we augment the synthesized sequences of reasoning steps for each sample in the training dataset by utilizing a state-of-the-art LLM for reasoning tasks. We then re-check the data based on the labels of each sample. We design two reward functions: one for language check, which uses a banned letter dictionary, and another for reasoning content, which employs Sentence Transformer models to measure the semantic similarity of the generated response compared to the corresponding reasoning data.

Our contributions are described as follows:
\begin{itemize}
    \item We propose algorithms and utilize our Vietnamese reasoning dataset to address the issue of language bias and ensure strict control over the reasoning content.
    \item We release a Vietnamese reasoning model with a medium size, specifically a 14.7 billion parameters, achieving a high overall accuracy of over 70\% on multiple-choice datasets, including the VLSP 2023 Challenge \cite{le2024overview} and SeaExam \cite{li2024seaexam}. 
    \item We also conduct experiments across multiple languages and demonstrate that reasoning-based answers significantly improve compared to few-shot learning techniques.
\end{itemize}

\section{Related Work}
\subsection{Chain-of-Thought}
Chain‑of‑Thought (CoT) prompting \cite{wei-2022} was introduced to encourage models to “think step by step”, providing a few exemplars with intermediate reasoning steps to improve multi‑step inference. Empirical results show that CoT significantly boosts performance on arithmetic, commonsense, and symbolic reasoning benchmarks, with a 540‑billion‑parameter model achieving state‑of‑the‑art accuracy on GSM8K using just eight CoT exemplars. Follow‑up work on self‑consistency decoding samples multiple reasoning paths and selects the most consistent answer, yielding substantial gains on \textsc{GSM8K} \(+17.9\%\), \textsc{SVAMP} (\(+11.0\%\)), \textsc{AQuA} (\(+12.2\%\)), \textsc{StrategyQA} (\(+6.4\%\)), and \textsc{ARC‑challenge} (\(+3.9\%\)) \cite{wang-2022}. These studies reveal that structured intermediate reasoning can be an emergent capability in sufficiently large models.

\subsection{Vietnamese Large Language Models}
While the domain of open-source models for the Vietnamese language is relatively nascent, there are already some notable models available. These include Vietcuna 3B\footnote{https://huggingface.co/vilm/vietcuna-3b}, Vietcuna-7B-v3\footnote{https://huggingface.co/vilm/vietcuna-7b-v3}, URA-LLaMA-7B\footnote{https://huggingface.co/ura-hcmut/ura-llama-7b}, and URA-LLaMA-13B\footnote{https://huggingface.co/ura-hcmut/ura-llama-13b}. Vietcuna-3B and Vietcuna-7B-v3 were developed from the foundational models BLOOMZ-3B\footnote{https://huggingface.co/bigscience/bloomz-3b} and BLOOMZ-7B1\footnote{https://huggingface.co/bigscience/bloomz-7b1} \cite{scao2022language}, respectively, and were further trained using 12GB of Vietnamese news texts for causal language modeling\footnote{https://www.vilm.org/research/how-did-we-train-vietcuna}. This process included fine-tuning with 200K instructional question and answer pairs, and 400K conversational samples. The URA-LLaMA models, originating from LLaMA-2, were pre-trained on Vietnamese content from Wikipedia and online news sources, with additional fine-tuning for instruction following. Furthermore, PhoGPT \cite{phogpt} have recently introduced the PhoGPT series, a new addition to the open-source generative models for Vietnamese, which includes a base 7.5B-parameter model and its instruction-following variant.

\subsection{Group Relative Policy Optimization}
Reinforcement learning (RL) is a subfield of Machine Learning (ML) in which an agent learns to make decisions through interactions with its environment, aiming to maximize cumulative rewards. When applied to LLMs, RL helps fine-tune these models to better align with human preferences and improve their performance on specialized tasks that require complex reasoning processes. A key category of RL algorithms is policy optimization, which focuses on directly refining the policy—the decision-making strategy an agent follows based on different states. GRPO was introduced in DeepSeekMath \cite{shao2402deepseekmath}, with the aim of improving the reasoning abilities of LLMs, especially in mathematical problem-solving and code generation. The reward serves as the foundation for the training signal, guiding the optimization direction in reinforcement learning. To train DeepSeek-R1-Zero \cite{guo2025deepseek}, authors implemented a rule-based reward mechanism comprising two primary reward types:
\begin{itemize}
    \item Format rewards: This function is used to evaluate the model's ability to generate responses that adhere to the desired structure.
    \item Accuracy rewards: This function is used to evaluate whether the extracted result (obtained from the response using a heuristic or structure-based method) matches the ground truth.
\end{itemize}

\section{Vietnamese Reasoning Dataset}
\subsection{Problem Definition}

\begin{figure}[!ht]
    \centering
    \includegraphics[width=1.0\linewidth]{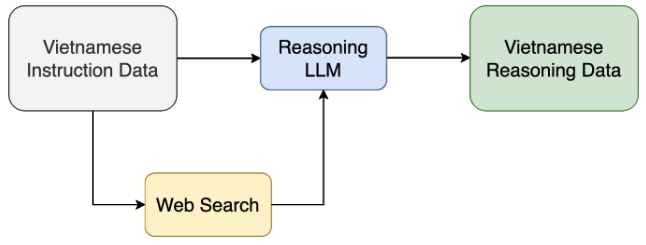}   
    \caption{Reasoning Data Curation}
    \label{fig:data_pipeline}
\end{figure}

In this section, we concentrate on curating high-quality Vietnamese reasoning tasks with verifiable answers. Each instance in the dataset consists of a pair of question-answer instruction $i \in I$ where $I$ represents the space of the instruction problems. The objective is to generate both a final answer $a \in A$ and a corresponding reasoning chain $r \in R$. We define a reasoning chain $r$ as a structured sequence of intermediate steps $\{s_1, s_2, ..., s_n\}$ where each step $s_{i}$ constitutes a logical deduction that incrementally bridges the initial question to the final answer. To enrich factual correctness and coverage, we also retrieve supplementary context $c \in C$ from the web using Google Search\footnote{https://google.com}. This additional context serves to enhance both the precision and depth of the model’s reasoning.

Formally, the reasoning process can be modeled as a function:

\hspace{2cm}$f: I \times C \to R \times A $

\subsection{Instruction Selection}
To ensure the robustness and generalizability of the reasoning capabilities of GreenMind, we design a multi-stage pipeline for selecting and curating instruction problems tailored for Vietnamese logical reasoning. Our selection process emphasizes linguistic diversity, logical depth, and cultural relevance. Specifically, we adopt the following criteria for instruction selection:

\begin{itemize} \item \textbf{Task Type Diversity:} We include a broad range of reasoning tasks such as arithmetic word problems, commonsense inference, symbolic logic, deductive and inductive reasoning, multi-hop question answering, and ethical dilemma evaluation. \item \textbf{Linguistic Complexity:} Instructions are sampled across varying syntactic and lexical complexities to challenge the model's understanding of nuanced Vietnamese expressions. \item \textbf{Reasoning Depth:} We prioritize tasks that require multi-step deductions, analogical thinking, and counterfactual reasoning over those solvable with shallow pattern matching. \item \textbf{Verifiability:} Each instruction-answer pair is manually verified or derived from trusted Vietnamese educational and encyclopedic sources, ensuring factual accuracy and clarity in logical steps. \end{itemize}
\subsection{Reasoning Chain Generation}
Beyond high-quality instructions, the generation of structured, verifiable reasoning chains is essential for training large language models capable of logical inference and multi-step deduction. To curate such high-quality solutions, we adopt a automated pipeline that incorporates web-scale retrieval to ensure factual correctness and logical coherence.

Given an instruction $i \in I$, we first retrieve supplementary context $c \in C$ from the web. This retrieved information often includes relevant definitions, background knowledge, or factual references that are not explicitly included in the instruction. The context $c$ serves as external knowledge to support the reasoning process, especially in tasks requiring factual grounding or domain-specific expertise.

Subsequently, we generate a reasoning chain $r = {s_1, s_2, ..., s_n}$, where each step $s_i$ is a logically valid and interpretable inference that incrementally bridges the gap between the given question and the final answer $a \in A$. These steps are structured to reflect a natural flow of thought, ensuring that the reasoning path remains traceable, coherent, and grounded in both the instruction and the retrieved context.

To ensure the quality of the generated reasoning chains, we apply a multi-stage filtering and validation process:

\begin{itemize} \item \textbf{Consistency Check:} We verify that the reasoning steps logically lead to the final answer and are internally consistent. \item \textbf{Redundancy Elimination:} Duplicate or unnecessary steps are pruned to maintain conciseness without sacrificing interpretability. \item \textbf{Format Conformity:} The reasoning chain must follow a step-by-step format to ensure compatibility with chain-of-thought (CoT) supervision. \end{itemize}

Moreover, to promote generalization and robustness, we also include examples with multiple valid reasoning chains for the same instruction. This encourages the model to develop a flexible reasoning strategy rather than memorizing fixed templates.

By focusing on both correctness and interpretability, our approach to reasoning chain generation enables GreenMind to demonstrate superior performance in structured reasoning tasks, setting a strong foundation for Vietnamese LLMs with transparent and explainable outputs

\section{GreenMind-Medium-14B-R1}
In this section, we present the base architecture, provide statistics on the Vietnamese training data, and describe the optimization strategy we used to transform the pretrained model into a Vietnamese-focused reasoning model.

\textbf{Base Model.} We utilize \texttt{Qwen2.5-14B-Instruct} \cite{qwen2.5} as a base model for finetuning process. \texttt{Qwen 2.5-14B-Instruct} is a dense, decoder-only Transformer language model comprising approximately 14.7 billion parameters. The architecture features 48 layers with a hidden state dimensionality of 5,120 and incorporates SwiGLU \cite{shazeer2020glu} feed-forward blocks alongside RMSNorm \cite{zhang2019root} normalization. The model employs Gated-Query Attention \cite{dhingra2016gated} (GQA) with 40 query heads and 8 key-value heads, augmented by Rotary Position Embeddings (RoPE) \cite{su2024roformer} combined with YaRN \cite{peng2309yarn} scaling to effectively support an extended context window of up to 128,000 tokens, enabling generation of sequences up to 8,192 tokens per request. Pre-training was conducted on an expansive multilingual corpus totaling 18 trillion tokens across more than 29 languages, including Vietnamese, representing a 2.5-fold increase over its version. Subsequently, the model underwent supervised fine-tuning on over one million high-quality instruction-response pairs, followed by staged reinforcement learning-based preference optimization. These design choices collectively enhance the model’s capacity for long-context understanding, multilingual comprehension, and instruction-following capabilities, making it well-suited for complex natural language processing tasks including code generation, mathematical reasoning, and structured data interpretation. This instruct model demonstrates strong, state-of-the-art performance across a range of academic and practical benchmarks, often outperforming models of similar or even larger sizes in several key domains, followed by their report \cite{qwen2.5}. 

\textbf{Training data.} We curated a high-quality Vietnamese instruction dataset with \textbf{55,418 samples}, each containing a question, a reasoning chain, and a final answer. To ensure broad generalization, instructions were drawn from diverse domains:
\begin{itemize} 
\item \textbf{Mathematics:} Mathematical problems train the model in symbolic reasoning, structured logic, and step-by-step problem solving, which are foundational for strong STEM-related performance. \cite{olmo-2024}. 
\item \textbf{Cultural:} These instructions cover Vietnamese idioms, proverbs, traditional practices, historical events, and literary references. This domain strengthens the model’s ability to interpret language with deep cultural semantics and regional specificity.
\item \textbf{Legal and Civic Knowledge:} Focused on basic legal concepts and civic education, particularly relevant in localized Vietnamese contexts such as laws, public policy, and social norms.
\item \textbf{Education and Exams:} Inspired by real-world school and university-level examination formats in Vietnam, fostering academic problem-solving patterns. 
\end{itemize}\
\begin{algorithm}[H]
\caption{Format}\label{algo:format}
    \begin{algorithmic}[1]
        \Require Completions $\mathbf{\mathcal{C}}$, regex of sequence format $rg_{s}$, regex of answer format $rg_{a}$, list of candidate results $l_{ans}$, score of completion structure $score_{c}$, score of answering structure $score_{a}$, score of answering candidate structure $score_{ac}$ 
        \Function{FORMAT-REWARDS}{$\mathbf{\mathcal{C}}$, $rg_{s}$, $rg_{a}$, $l_{ans}$, $score_c$, $score_a$, $score_{ac}$}
            \State $l_{scores} = []$ \Comment{List of scores}
            \For{$i = 1$ to $length(\mathbf{\mathcal{C}})$}
                \State $score \gets 1.0$
                \If{$not\_match(\mathbf{\mathcal{C}}_{i}, rg_{s})$}
                    \State $score \gets score - score_{c}$
                \EndIf
                \State $\hat{p} \gets find(\mathbf{\mathcal{C}}_{i}, rg_{a})$ \Comment{Get predictions}
                \If{$length(\hat{p}) == 1$}
                    \If{$\hat{p} \nsubseteq l_{ans}$}
                        \State $score \gets score - score_{ac}$
                    \EndIf
                \Else
                    \State $score \gets score - score_{a}$
                \EndIf
                \State Append $score$ to $l_{scores}$
            \EndFor
            \State \Return $l_{scores}$
        \EndFunction
    \end{algorithmic}
\end{algorithm}
\textbf{Optimization with reward functions.}
We fine-tune the model to focus on tasks that require generating concise answers, which involve a step-by-step reasoning process. Following DeepSeek-R1 \cite{guo2025deepseek}, we design two fundamental reward functions.
\begin{itemize}
    \item Format rewards: Our objective is to ensure that reasoning chains are enclosed within the \texttt{<think>...</think>} tags, and that the final answer is enclosed within the \texttt{<answer>...</answer>} tags. Among these, we place greater emphasis on the structure of the final answer, as it remains the ultimate goal to be achieved. Details on how the format reward is computed are described in Algorithm~\ref{algo:format}. To enable smooth reward assignment, we recommend that:
    \[
    \begin{cases}
    0.0 < \text{score}_{ac} < \text{score}_a, \text{score}_c < 1.0\\
    \text{score}_c + \text{score}_a = 1.0 \\
    \end{cases}
    \]
    \item Accuracy rewards: This function is used to assign scores to the generated answers. A prerequisite is that the prediction must be extracted from the completion based on the expected answering structure. For tasks involving multiple choices, we still encourage assigning partial score to help the model capture the scope of possible outcomes, e.g.\ one of candidates $[A, B, C, D, E]$ (see Algorithm~\ref{algo:answering}).
\end{itemize}

To the best of our knowledge, there is no specific statistical report on language data distribution mentioned in the papers or technical reports by the authors of the Qwen model family. Therefore, we identify potential biases toward specific languages through empirical observations. The results for base model indicate that the sampling process can effectively handle Vietnamese language, although Chinese characters occasionally appear. Additionally, the reasoning content should be tightly controlled and remain within the scope of the original topic to ensure alignment with the final answer or the list of possible answers. In summary, we propose the design of two additional reward functions to address the above challenges:
\begin{itemize}
    \item Language rewards: We define a banned character list to penalize undesired language usage. Our goal is to guide the model to generate content in a single language that matches the input query. Therefore, a completion receives a reward if and only if it does not contain any characters from the banned list (see Algorithm~\ref{algo:language}).
    \item Semantic similarity rewards: Based on our proposed Vietnamese reasoning dataset, we measure the closeness of completions using a Sentence Transformers-based model. The selection model should be validated to ensure good performance on the specific monolingual setting. With Algorithm~\ref{algo:reasoning}, the cosine score is preserved if it exceeds a predefined threshold; otherwise, it is set to zero to mitigate the risk of hallucination.
\end{itemize}
\begin{algorithm}
\caption{Answering}\label{algo:answering}
    \begin{algorithmic}[1]
        \Require Completions $\mathbf{\mathcal{C}}$, regex of answering format $rg_{a}$, list of candidate results $l_{ans}$, score of answering candidates $score_{ac}$, ground truth $gt$ 
        \Function{ANSWERING-REWARDS}{$\mathbf{\mathcal{C}}$, $rg_{a}$, $l_{ans}$, $score_{ac}$, $gt$ }
            \State $l_{scores} = []$ \Comment{List of scores}
            \For{$i = 1$ to $length(\mathbf{\mathcal{C}})$}
                \State $score \gets 0.0$
                \State $\hat{p} \gets find(\mathbf{\mathcal{C}}_{i}, rg_{a})$ \Comment{Get predictions}
                \If{$length(\hat{p}) == 1$}
                    \If{$\hat{p} == gt$}
                        \State $score \gets 1$
                    \ElsIf{$\hat{p} \subset l_{ans}$}
                        \State $score \gets score_{ac}$ \Comment{For multiple-choices tasks}
                    \Else
                        \State $score \gets 0$
                    \EndIf
                \Else
                    \State $score \gets 0$
                \EndIf
                \State Append $score$ to $l_{scores}$
            \EndFor
            \State \Return $l_{scores}$
        \EndFunction
    \end{algorithmic}
\end{algorithm}
\begin{algorithm}
\caption{Language}\label{algo:language}
    \begin{algorithmic}[1]
        \Require Completions $\mathbf{\mathcal{C}}$, List of banned letters $l_{bl}$ 
        \Function{LANGUAGE-REWARDS}{$\mathbf{\mathcal{C}}$, $l_{bl}$}
            \State $l_{scores} = []$ \Comment{List of scores}
            \For{$i = 1$ to $length(\mathbf{\mathcal{C}})$}
                \State $score \gets 1.0$
                \If{$exist(\mathbf{c\in\mathcal{C}_i}, l_{bl})$}
                    \State $score \gets 0.0$
                \EndIf
                \State Append $score$ to $l_{scores}$
            \EndFor
            \State \Return $l_{scores}$
        \EndFunction
    \end{algorithmic}
\end{algorithm}
\section{Experiment}
\subsection{Implementation Details}
We fine-tune our model with all parameters (full fine-tuning) to ensure optimal performance, allowing the model to fully adapt to the downstream task and leverage the capacity of all layers for improved generalization. We use DeepSpeed \cite{rajbhandari2020zero} framework to enable efficient large-scale model training by reducing memory footprint and accelerating training throughput. Specifically, we leverage ZeRO Stage 3 to partition optimizer states, gradients, and parameters across GPUs, which allows us to train models that would otherwise exceed device memory limitations. Additionally, mixed-precision training further improves computational efficiency without sacrificing model accuracy. Additionally, we report the configuration of the hyperparameters used during the fine-tuning process, as detailed in Table~\ref{tab:training-hyperparams}. 
\begin{algorithm}[H]
\caption{Semantic Similarity of Reasoning Content}\label{algo:reasoning}
    \begin{algorithmic}[1]
        \Require Completions $\mathbf{\mathcal{C}}$, Sentence Transformers model $ST\_{model}$, reasoning data $rs$, similarity threshold $\xi$
        \Function{SS-REWARDS}{$\mathbf{\mathcal{C}}$, $ST\_{model}$, $rs$, $\xi$}
            \State $l_{scores} = []$ \Comment{List of scores}
            \For{$i = 1$ to $length(\mathbf{\mathcal{C}})$}
                \State $score \gets ST\_{model}(\mathcal{C}_i, rs_i)$
                \If{$score < \xi$}
                    \State $score \gets 0.0$
                \EndIf                
                \State Append $score$ to $l_{scores}$
            \EndFor
            \State \Return $l_{scores}$
        \EndFunction
    \end{algorithmic}
\end{algorithm}
We utilize 7 GPUs for model fine-tuning and, one GPU for inference by employing the vLLM framework \cite{kwon2023efficient}. Our objective is to produce completions that are creative while mitigating hallucinations. The hyperparameters for inferencing are presented in the Table~\ref{tab:inference-hyperparams}. All experiments were conducted on 8 H100 GPUs.

\begin{table}[h!]
\centering
\caption{Training Hyperparameters}
\resizebox{0.7\linewidth}{!}{ 
\begin{tabular}{ll}
\toprule
\textbf{Hyperparameter} & \textbf{Value} \\
\midrule
epochs & 4 \\
per\_device\_train\_batch\_size & 1 \\
gradient\_accumulation\_steps & 8 \\
gradient\_checkpointing & true \\
learning\_rate & 5.0e-7 \\
lr\_scheduler\_type & cosine \\
warmup\_ratio & 0.03 \\
beta & 0.001 \\
max\_prompt\_length & 256 \\
max\_completion\_length & 1024 \\
num\_generations & 4 \\
use\_vllm & true \\
vllm\_gpu\_memory\_utilization & 0.9 \\
\bottomrule
\end{tabular}
}
\label{tab:training-hyperparams}
\end{table}

\begin{table}[h!]
\centering
\caption{vLLM Inferencing Hyperparameters}
\resizebox{0.5\linewidth}{!}{ 
\begin{tabular}{ll}
\toprule
\textbf{Hyperparameter} & \textbf{Value} \\
\midrule
Repetition Penalty & 1.2 \\
Temperature        & 0.6 \\
Top-p (nucleus)    & 0.8 \\
Top-k              & 4   \\
\bottomrule
\end{tabular}
}
\label{tab:inference-hyperparams}
\end{table}

\begin{table}[ht]
\centering
\caption{SeaExam performance compared to SOTA model}
\resizebox{1.1\linewidth}{!}{ 
\begin{tabular}{lcccc}
\toprule
\textbf{Model} & \textbf{SeaExam-ID} & \textbf{SeaExam-TH} & \textbf{SeaExam-VI} & \textbf{Avg} \\
\midrule
Meta-Llama-3.1-70B-Instruct & 65.8 & \textbf{70.6} & 72.6 & 69.7 \\
gemma3-27b-it & 64.4 & 67.5 & 73.1 & 68.4 \\
Qwen2.5-14B-Instruct & 67.6 & 68.8 & 73.1 & 69.8 \\
GreenMind-Medium-14B-R1 & \textbf{74.36} & 69.75 & \textbf{74.44} & \textbf{72.79} \\
\bottomrule
\end{tabular}
}
\label{tab:seaexam-performance-public}
\end{table}

\begin{table*}[ht]
\centering
\resizebox{\textwidth}{!}{%
\begin{tabular}{lcccccc}
\toprule
\textbf{Model} & \textbf{Access} & \textbf{STEM} & \textbf{Social Science} & \textbf{Humanities} & \textbf{Others} & \textbf{Avg} \\
\midrule

VNPTAI.IO-Medium-R1 & Private & 77.09 & 82.3 & 78.85 & 69.98 & 77.43 \\
MISA-Llama3-v1.1 & Private & 77.5 & 80.75 & 76.62 & 71.6 & 76.87 \\
BnK-AI-Medium-v2  & Private & 80.94 & 80.76 & 70.7 & 74.06 & 76.66 \\
VNPTAI.IO-Large-v4 & Private & 78.05 & 79.05 & 75.39 & 70.37 & 76.21 \\
GreenNode-xMedium-v1 & Private & 75.7 & 81.09 & 75.25 & 69.33 & 75.5 \\
GreenMind-Medium-14B-R1 (Ours) & Weight & 76.78 & 77.36 & 72.32 & 69.03 & 74.29 \\
CakebyVPBank-Large & Private & 77.75 & 78.11 & 70.38 & 67.82 & 73.99 \\
DeepSeek-R1-Distill-Llama-70B & Weight & 76.77 & 76.23 & 67.98 & 66.82 & 72.41 \\

\bottomrule
\end{tabular}
}
\caption{VMLU performance compared to fine-tuned models}
\label{tab:vmlu-performance-public}
\end{table*}

\begin{table*}[ht]
\centering
\resizebox{\textwidth}{!}{%
\begin{tabular}{lccccc}
\toprule
\textbf{Model} & \textbf{ComprehensionQA-vi} $\uparrow$ & \textbf{Exams-vi} $\uparrow$ & \textbf{LAMBADA-vi} $\downarrow$ & \textbf{WikiQA-vi} $\uparrow$ & \textbf{MMLU-vi} $\uparrow$ \\
\midrule
cpt-smartbot-13b & 0.6633 & 0.3473 & 21.9864 & 0.4455 & 0.414 \\
ura-llama-13b & 0.6556 & 0.342 & 17.5614 & 0.438 & 0.3973 \\
greennode-7b (prior work) & 0.6122 & 0.2892 & 189.7782 & 0.3335 & 0.387 \\
greennode-14b (prior work) & 0.6711 & 0.3672 & 29.5967 & 0.468 & 0.5281 \\
GreenMind-Medium-14B-R1 (our) & \textbf{0.8689} & \textbf{0.7796} & \textbf{10.7609} & \textbf{0.7915} & \textbf{0.7124} \\
\bottomrule
\end{tabular}
}
\caption{\textbf{VLSP 2023 Challenge.} The performance of our model outperforms most SOTA models.}
\label{tab:vlsp2023-performance-public}
\end{table*}

\subsection{Experimental Results}
\begin{figure}
    \centering
    \includegraphics[width=1.0\linewidth]{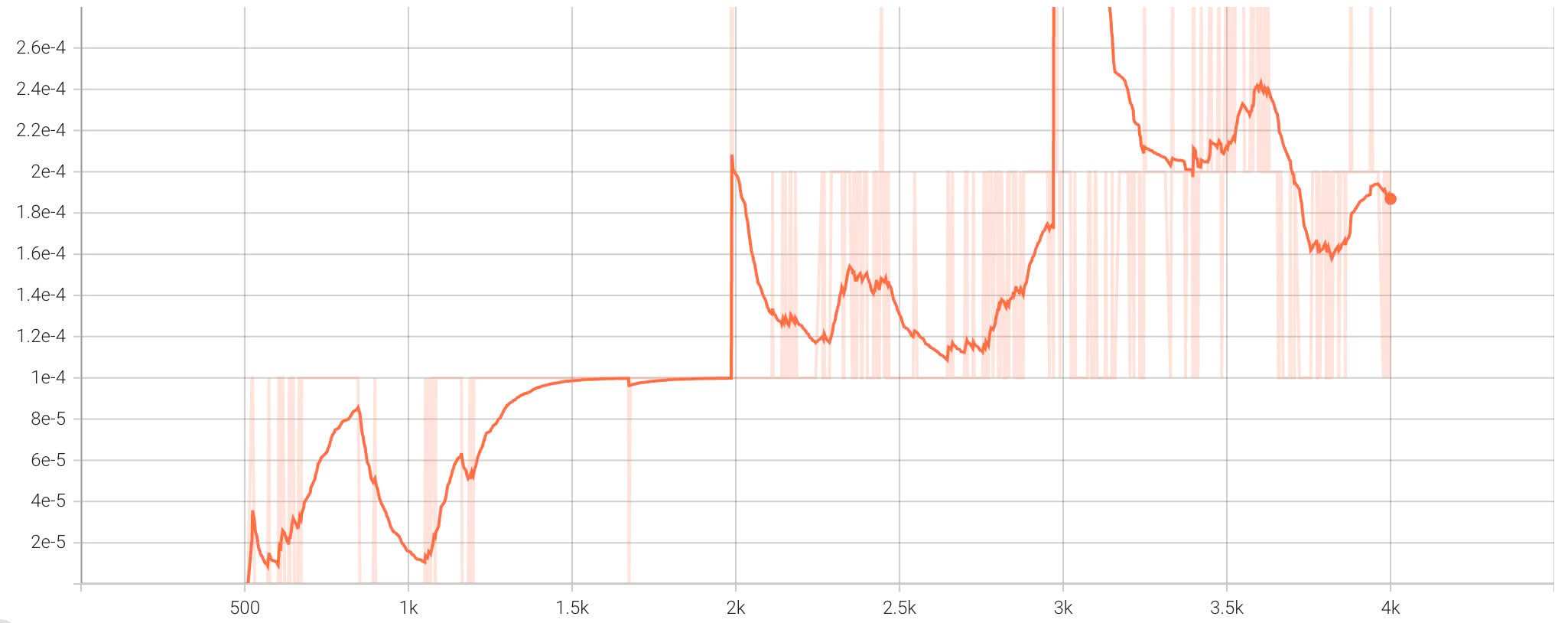}
    \caption{Training Loss.}
    \label{fig:train-loss}
\end{figure}

\textbf{Finetuning results.} We present some basic analysis after fine-tuning for approximately 4 epochs, as reported in Figure~\ref{fig:train-loss}. The GRPO loss function starts at 0 and gradually increases. The reason is that the Kullback-Leibler divergence approaches infinity as the distributions of $\pi_{\theta}$ and $\pi_{ref}$ become more different.

\textbf{Quantitative Evaluation.} Experimental results on the SeaExam multiple-choice dataset \cite{li2024seaexam} show that our reasoning model outperforms most Southeast Asian languages, as well as the overall average across all languages, when compared to baseline models with significantly larger parameter sizes — including those with up to 70 billion parameters (see Table~\ref{tab:seaexam-performance-public}). Notably, these models were evaluated under few-shot prompting settings. On the VLSP 2023 Challenge dataset \cite{le2024overview}, our model achieves superior performance over all previously reported models. In particular, \texttt{greennode-7b} and \texttt{greennode-14b} were trained using Supervised Fine-Tuning (SFT) and Direct Preference Optimization (DPO) \cite{rafailov2023direct}, respectively. Regarding the VMLU\footnote{\url{https://vmlu.ai}} dataset, at the current time, the only accessible model is \texttt{DeepSeek-R1-Distill-Llama-70B}\footnote{https://huggingface.co/deepseek-ai/DeepSeek-R1-Distill-Llama-70B}. Results indicate that our model performs slightly better across most topics, including both mathematical and social science domains. 

\textbf{Qualitative results.} The qualitative results are presented in the Table~\ref{tab:generate}. These results demonstrate that the model's responses adhere to structural rules of reasoning and outcome formulation. We showcase reasoning chains across various topics, ranging from natural sciences to social sciences. The visualizations illustrate that the model performs a sequence of logical inferences before arriving at the final answer.

\begin{table*}[!h]
\begin{center}
\begin{tabularx}{\textwidth}{l X} 
\hline
\\
\textbf{Task:} & \textbf{Mathematics}\\
\hline
\textbf{Prompt:} & Hình lập phương có cạnh là 5 m. Vậy thể tích hình đó là: 

A. $150 m^{3}$

B. $25 m^{3}$

C. $100 m^{3}$

D. $125 m^{3}$ \\
\hline
\textbf{Answer:} & <think>Đầu tiên, chúng ta cần nhớ công thức tính thể tích của hình lập phương:
Thể tích = a³ (trong đó a là độ dài cạnh của hình lập phương)
Trong trường hợp này, cạnh của hình lập phương là 5m.

Tiếp theo, áp dụng công thức vào dữ liệu cho sẵn:
V = 5³ = 5 × 5 × 5 = 125 m³

Cuối cùng, so sánh kết quả với các lựa chọn được cung cấp:
- A. 150 m³
- B. 25 m³
- C. 100 m³
- D. 125 m³

Kết quả đúng nhất chính là 125 m³.</think>
<answer>D</answer> \\
\hline
\\

\textbf{Task:} & \textbf{Natural Science}\\
\hline
\textbf{Prompt:} & Các cơ quan sinh dục cái và cơ quan sinh dục đực của thực vật có hoa là: 

A. Đài hoa và nhụy hoa.

B. Cánh hoa và bao phấn.

C. Nhụy và nhị.

D. Cả 3 ý trên. \\
\hline
\textbf{Answer:} & <think>Đầu tiên, chúng ta cần xác định:

- Cơ quan sinh dục cái (nữ) ở thực vật có hoa là gì?

- Cơ quan sinh dục đực (nam) ở thực vật có hoa là gì?

1. Cơ quan sinh dục cái thường gồm phần noãn hay kén noãn nằm bên dưới.
2. Cơ quan sinh dục đực thường gồm nhị hoa sản xuất tinh trùng dạng phấn hoa.

Tiếp theo, phân tích mỗi phương án:

A - Đài hoa và nhụy hoa không chính xác vì đây chỉ là bộ phận bảo vệ hoặc phụ trợ.

B - Cánh hoa và bao phấn cũng tương tự như trên.

C - Nhụy và nhị đúng với mô tả phía trên.

D - Tất cả ba đều sai khi xét đến vị trí cụ thể của cơ quan sinh dục.

Vậy rõ ràng nhất từ góc nhìn cấu trúc chức năng thì C sẽ phù hợp.</think>
<answer>C</answer>\\
\hline
\\
\textbf{Task:} & \textbf{Social Science}\\
\hline
\textbf{Prompt:} & Chống bão ở nước ta phải luôn kết hợp với chống

A. hạn hán.

B. động đất.

C. ngập lụt.

D. sương muối. \\
\hline
\textbf{Answer:} & <think>Đầu tiên, chúng cần xác định vấn đề chính của câu hỏi: "chống bão".
Tiếp theo, phân tích mỗi phương án:

- Phương án A: Hạn hán - Đây không liên quan trực tiếp đến hậu quả thường gặp khi có bão.

- Phương án B: Động đất - Cũng không liên quan nhiều tới tình hình thời tiết do bão gây nên.

- Phương án C: Ngập lụt - Khi xảy ra bão, mưa lớn dẫn đến nguy cơ ngập lụt rất cao.

- Phương án D: Sương muối - Thường xuất hiện vào mùa khô hoặc trên vùng biển nhất định, ít liên quan đến bão.
Từ đó, rõ ràng phương án phù hợp nhất sẽ là sự kết hợp giữa chống bão và chống ngập lụt vì hai yếu tố này đi cùng nhau trong thực tế.</think>
<answer>C</answer>\\
\hline
\end{tabularx}
\end{center}
\caption{Qualitative results on SeaExam with \texttt{GreenMind-Medium-14B-R1}.}
\label{tab:generate}
\end{table*}

\section{Conclusion} 
We release \textbf{GreenMind-Medium-14B-R1}, a medium-sized Vietnamese language model capable of effectively addressing questions that require intermediate-level reasoning, such as general knowledge and social science topics. By leveraging the GRPO strategy for fine-tuning, we guide the model to generate logically coherent responses. This approach aims to provide users with informative answers, as well as intuitive explanations—valuable not only for end users but also for further research in improving data quality and sampling techniques.
\section*{Acknowledgments}
We sincerely express our deep appreciation to \textbf{GreenNode.ai}\footnote{\url{https://greennode.ai/}}, our affiliated organization, for their unwavering support throughout the course of this research. GreenNode.ai has played a pivotal role by providing essential resources—most notably, access to high-performance H100 GPUs—which significantly accelerated the fine-tuning process of our models. This generous support was instrumental in the successful development of a Vietnamese reasoning language model.

\bibliography{ref}

\begin{thebibliography}{20}
\expandafter\ifx\csname natexlab\endcsname\relax\def\natexlab#1{#1}\fi

\bibitem[{Dhingra et~al.(2016)Dhingra, Liu, Yang, Cohen, and Salakhutdinov}]{dhingra2016gated}
Bhuwan Dhingra, Hanxiao Liu, Zhilin Yang, William~W Cohen, and Ruslan Salakhutdinov. 2016.
\newblock Gated-attention readers for text comprehension.
\newblock \emph{arXiv preprint arXiv:1606.01549}.

\bibitem[{Guo et~al.(2025)Guo, Yang, Zhang, Song, Zhang, Xu, Zhu, Ma, Wang, Bi et~al.}]{guo2025deepseek}
Daya Guo, Dejian Yang, Haowei Zhang, Junxiao Song, Ruoyu Zhang, Runxin Xu, Qihao Zhu, Shirong Ma, Peiyi Wang, Xiao Bi, et~al. 2025.
\newblock Deepseek-r1: Incentivizing reasoning capability in llms via reinforcement learning.
\newblock \emph{arXiv preprint arXiv:2501.12948}.

\bibitem[{Kwon et~al.(2023)Kwon, Li, Zhuang, Sheng, Zheng, Yu, Gonzalez, Zhang, and Stoica}]{kwon2023efficient}
Woosuk Kwon, Zhuohan Li, Siyuan Zhuang, Ying Sheng, Lianmin Zheng, Cody~Hao Yu, Joseph~E. Gonzalez, Hao Zhang, and Ion Stoica. 2023.
\newblock Efficient memory management for large language model serving with pagedattention.
\newblock In \emph{Proceedings of the ACM SIGOPS 29th Symposium on Operating Systems Principles}.

\bibitem[{Le et~al.(2024)Le, Can, Nguyen, and Tran}]{le2024overview}
Hoang-Quynh Le, Duy-Cat Can, Khanh-Vinh Nguyen, and Mai-Vu Tran. 2024.
\newblock \href {https://arxiv.org/abs/2402.13613} {Overview of the vlsp 2023 -- comom shared task: A data challenge for comparative opinion mining from vietnamese product reviews}.
\newblock \emph{arXiv preprint arXiv:2402.13613}.

\bibitem[{Li et~al.(2024)Li, Tan, Wang, Zhang, Wang, Wang, Liu, Wang, Liu, and Liu}]{li2024seaexam}
Yixuan Li, Xu~Tan, Yichong Wang, Zihan Zhang, Longyue Wang, Shuo Wang, Xiaohua Liu, Rui Wang, Jingjing Liu, and Tie-Yan Liu. 2024.
\newblock \href {https://arxiv.org/abs/2404.11086} {Seaexam: Benchmarking large language models for southeast asian languages with human exam questions}.
\newblock \emph{arXiv preprint arXiv:2404.11086}.

\bibitem[{Nguyen et~al.(2023)Nguyen, Nguyen, Tran, Nguyen, Nguyen, Nguyen, Phung, and Bui}]{phogpt}
Dat~Quoc Nguyen, Linh~The Nguyen, Chi Tran, Dung~Ngoc Nguyen, Nhung Nguyen, Thien~Huu Nguyen, Dinh Phung, and Hung Bui. 2023.
\newblock {PhoGPT: Generative Pre-training for Vietnamese}.
\newblock \emph{arXiv preprint}, arXiv:2311.02945.

\bibitem[{OLMo et~al.(2024)OLMo, Walsh, Soldaini, Groeneveld, Lo, Arora, Bhagia, Gu, Huang, Jordan, Lambert, Schwenk, Tafjord, Anderson, Atkinson, Brahman, Clark, Dasigi, Dziri, Guerquin, Ivison, Koh, Liu, Malik, Merrill, Miranda, Morrison, Murray, Nam, Pyatkin, Rangapur, Schmitz, Skjonsberg, Wadden, Wilhelm, Wilson, Zettlemoyer, Farhadi, Smith, and Hajishirzi}]{olmo-2024}
Team OLMo, Pete Walsh, Luca Soldaini, Dirk Groeneveld, Kyle Lo, Shane Arora, Akshita Bhagia, Yuling Gu, Shengyi Huang, Matt Jordan, Nathan Lambert, Dustin Schwenk, Oyvind Tafjord, Taira Anderson, David Atkinson, Faeze Brahman, Christopher Clark, Pradeep Dasigi, Nouha Dziri, Michal Guerquin, Hamish Ivison, Pang~Wei Koh, Jiacheng Liu, Saumya Malik, William Merrill, Lester~James Miranda, V, Jacob Morrison, Tyler Murray, Crystal Nam, Valentina Pyatkin, Aman Rangapur, Michael Schmitz, Sam Skjonsberg, David Wadden, Christopher Wilhelm, Michael Wilson, Luke Zettlemoyer, Ali Farhadi, Noah~A. Smith, and Hannaneh Hajishirzi. 2024.
\newblock \href {https://doi.org/10.48550/arxiv.2501.00656} {{2 OLMO 2 Furious}}.
\newblock \emph{arXiv (Cornell University)}.

\bibitem[{Ouyang et~al.(2022)Ouyang, Wu, Jiang, Almeida, Wainwright, Mishkin, Zhang, Agarwal, Slama, Ray et~al.}]{ouyang2022training}
Long Ouyang, Jeffrey Wu, Xu~Jiang, Diogo Almeida, Carroll Wainwright, Pamela Mishkin, Chong Zhang, Sandhini Agarwal, Katarina Slama, Alex Ray, et~al. 2022.
\newblock Training language models to follow instructions with human feedback.
\newblock \emph{Advances in neural information processing systems}, 35:27730--27744.

\bibitem[{Peng et~al.()Peng, Quesnelle, Fan, and Shippole}]{peng2309yarn}
Bowen Peng, Jeffrey Quesnelle, Honglu Fan, and Enrico Shippole.
\newblock Yarn: Efficient context window extension of large language models, 2023.
\newblock \emph{URL https://arxiv. org/abs/2309.00071}.

\bibitem[{Rafailov et~al.(2023)Rafailov, Sharma, Mitchell, Manning, Ermon, and Finn}]{rafailov2023direct}
Rafael Rafailov, Archit Sharma, Eric Mitchell, Christopher~D Manning, Stefano Ermon, and Chelsea Finn. 2023.
\newblock Direct preference optimization: Your language model is secretly a reward model.
\newblock \emph{Advances in Neural Information Processing Systems}, 36:53728--53741.

\bibitem[{Rajbhandari et~al.(2020)Rajbhandari, Rasley, Ruwase, and He}]{rajbhandari2020zero}
Samyam Rajbhandari, Jeff Rasley, Olatunji Ruwase, and Yuxiong He. 2020.
\newblock Zero: Memory optimizations toward training trillion parameter models.
\newblock In \emph{SC20: International Conference for High Performance Computing, Networking, Storage and Analysis}, pages 1--16. IEEE.

\bibitem[{Scao et~al.(2022)Scao, Wang, Hesslow, Saulnier, Bekman, Bari, Biderman, Elsahar, Muennighoff, Phang et~al.}]{scao2022language}
Teven~Le Scao, Thomas Wang, Daniel Hesslow, Lucile Saulnier, Stas Bekman, M~Saiful Bari, Stella Biderman, Hady Elsahar, Niklas Muennighoff, Jason Phang, et~al. 2022.
\newblock What language model to train if you have one million gpu hours?
\newblock \emph{arXiv preprint arXiv:2210.15424}.

\bibitem[{Shao et~al.()Shao, Wang, Zhu, Xu, Song, Bi, Zhang, Zhang, Li, Wu et~al.}]{shao2402deepseekmath}
Zhihong Shao, Peiyi Wang, Qihao Zhu, Runxin Xu, Junxiao Song, Xiao Bi, Haowei Zhang, Mingchuan Zhang, YK~Li, Y~Wu, et~al.
\newblock Deepseekmath: Pushing the limits of mathematical reasoning in open language models, 2024.
\newblock \emph{URL https://arxiv. org/abs/2402.03300}.

\bibitem[{Shazeer(2020)}]{shazeer2020glu}
Noam Shazeer. 2020.
\newblock Glu variants improve transformer.
\newblock \emph{arXiv preprint arXiv:2002.05202}.

\bibitem[{Su et~al.(2024)Su, Ahmed, Lu, Pan, Bo, and Liu}]{su2024roformer}
Jianlin Su, Murtadha Ahmed, Yu~Lu, Shengfeng Pan, Wen Bo, and Yunfeng Liu. 2024.
\newblock Roformer: Enhanced transformer with rotary position embedding.
\newblock \emph{Neurocomputing}, 568:127063.

\bibitem[{Team(2024)}]{qwen2.5}
Qwen Team. 2024.
\newblock \href {https://qwenlm.github.io/blog/qwen2.5/} {Qwen2.5: A party of foundation models}.

\bibitem[{Wang et~al.(2022)Wang, Wei, Schuurmans, Le, Chi, and Zhou}]{wang-2022}
Xuezhi Wang, Jason Wei, Dale Schuurmans, Quoc Le, Ed~Chi, and Denny Zhou. 2022.
\newblock \href {https://doi.org/10.48550/arxiv.2203.11171} {{Self-Consistency improves chain of thought reasoning in language models}}.
\newblock \emph{arXiv (Cornell University)}.

\bibitem[{Wei et~al.(2022{\natexlab{a}})Wei, Wang, Schuurmans, Bosma, Chi, Le, and Zhou}]{wei-2022}
Jason Wei, Xuezhi Wang, Dale Schuurmans, Maarten Bosma, Ed~Chi, Quoc Le, and Denny Zhou. 2022{\natexlab{a}}.
\newblock \href {https://doi.org/10.48550/arxiv.2201.11903} {{Chain-of-Thought prompting elicits reasoning in large language models}}.
\newblock \emph{arXiv (Cornell University)}.

\bibitem[{Wei et~al.(2022{\natexlab{b}})Wei, Wang, Schuurmans, Bosma, Xia, Chi, Le, Zhou et~al.}]{wei2022chain}
Jason Wei, Xuezhi Wang, Dale Schuurmans, Maarten Bosma, Fei Xia, Ed~Chi, Quoc~V Le, Denny Zhou, et~al. 2022{\natexlab{b}}.
\newblock Chain-of-thought prompting elicits reasoning in large language models.
\newblock \emph{Advances in neural information processing systems}, 35:24824--24837.

\bibitem[{Zhang and Sennrich(2019)}]{zhang2019root}
Biao Zhang and Rico Sennrich. 2019.
\newblock Root mean square layer normalization.
\newblock \emph{Advances in Neural Information Processing Systems}, 32.

\end{thebibliography}
\bibliographystyle{acl_natbib}

\end{document}